# Ein semi-automatischer graphbasierter Ansatz zur Bestimmung des Randes von eloquenten Faserverbindungen des Gehirns
# A Semi-Automatic Graph-Based Approach for Determining the Boundary of Eloquent Fiber Bundles in the Human Brain

M. H .A. Bauer[1,2], J. Egger[1,2], D. Kuhnt[1], S. Barbieri[3], J. Klein[3], H.-K. Hahn[3], B. Freisleben[2], Ch. Nimsky[1]
[1]Universität Marburg, Klinik für Neurochirurgie, Marburg, Germany
[2]Universität Marburg, Fachbereich Mathematik und Informatik, Marburg, Germany
[3]Fraunhofer MEVIS, Bremen, Germany

bauermi@med.uni-marburg.de

## Kurzfassung

Mit Hilfe der Diffusions-Tensor-Bildgebung (DTI) ist es möglich die Lage, Orientierung und Ausdehnung von Bahnsystemen der weißen Substanz im menschlichen Gehirn zu schätzen. Das nicht-invasive bildgebende Verfahren macht sich hierbei die Diffusion von Wassermolekülen zunutze und bestimmt für jedes Volumenelement (Voxel) die nötigen Diffusionskoeffizienten. Die Bestimmung dieser Diffusionskoeffizienten und die dadurch mögliche Ableitung von Informationen über Faserbahnen ist von besonderem Interesse bei der Planung und Durchführung neurochirurgischer Eingriffe. Um das Risiko neuronaler Defizite bei Operationen im Kopfbereich – wie beispielsweise der Resektion von Tumoren (Gliomen) – für den Patienten zu minimieren ist die Segmentierung und Integration der Ergebnisse in den OP-Situs von größter Bedeutung. Im Rahmen dieses Beitrages wird ein robuster und effizienter graph-basierter Ansatz zur Segmentierung röhrenförmiger Faserbahnen im menschlichen Gehirn vorgestellt. Zur Definition der Kostenfunktion wird die fraktionelle Anisotropie (FA) genutzt, die sich in den Daten jedoch von Patient zu Patient unterscheiden kann. Neben der manuellen Definition der Saatregionen war daher bis jetzt auch immer zusätzlich eine manuelle Definition der Kostenfunktion durch den Benutzer notwendig. Um den Ansatz weiter zu automatisieren wird im vorliegenden Beitrag eine Lösung zur automatischen Bestimmung der manuellen Kostenfunktion durch den Einsatz verschiedener 3D Masken in den individuellen Daten vorgestellt.

## Abstract

Diffusion Tensor Imaging (DTI) allows estimating the position, orientation and dimension of bundles of nerve pathways. This non-invasive imaging technique takes advantage of the diffusion of water molecules and determines the diffusion coefficients for every voxel of the data set. The identification of the diffusion coefficients and the derivation of information about fiber bundles is of major interest for planning and performing neurosurgical interventions. To minimize the risk of neural deficits during brain surgery as tumor resection (e.g. glioma), the segmentation and integration of the results in the operating room is of prime importance. In this contribution, a robust and efficient graph-based approach for segmentating tubular fiber bundles in the human brain is presented. To define a cost function, the fractional anisotropy (FA) is used, derived from the DTI data, but this value may differ from patient to patient. Besides manually defining seed regions describing the structure of interest, additionally a manual definition of the cost function by the user is necessary. To improve the approach the contribution introduces a solution for automatically determining the cost function by using different 3D masks for each individual data set.

## 1 Einleitung

Diffusions-Tensor Bildgebung (DTI) als nicht-invasives bildgebendes Verfahren bietet Informationen über die Lage und den Verlauf von Bahnsystemen der weißen Substanz im menschlichen Gehirn auf Basis der Diffusion von Wassermolekülen. Es gibt verschiedene Segmentierungsverfahren, wie z. B. Fiber Tracking, mit denen der Verlauf bestimmt werden kann. Dabei ist die Grenze der eloquenten Struktur im Kontext neurochirurgischer Operationen wie zum Beispiel der Resektion von Tumoren in unmittelbarer Nachbarschaft zu wichtigen Faserbahnen wie der Pyramidenbahn (Motorik) oder der Sehbahnen unerlässlich. Zur Integration in die OP-Planung müssen daher zunächst die eloquenten Strukturen segmentiert werden und das resultierende 3D-Objekt bzw. die zugehörigen 2D-Ansichten in die Navigation integriert werden.

In diesem Beitrag wird ein neuer Ansatz zur semi-automatischen Segmentierung von eloquenten Faserverbindungen im menschlichen Gehirn vorgestellt. Ausgehend von manuell definierten Saatregionen wird ein gerichteter, gewichteter 3D-Graph mit automatisch definierter Kostenfunktion aufgebaut. Durch einen minimalen s-t-Schnitt wird das Faserbündel vom Hintergrund optimal getrennt. Evaluiert wurde der Ansatz mit verschiedenen Software-Phantomen.

Der vorliegende Beitrag ist wie folgt aufgebaut: In Abschnitt 2 wird der Stand der Forschung dargestellt. Abschnitt 3 beinhaltet den vorgestellten Ansatz. In Abschnitt 4 folgt die Diskussion der experimentellen Ergebnisse. Abschließend fasst Abschnitt 5 den Beitrag zusammen und gibt einen Ausblick über zukünftige Erweiterungen.

## 2 Stand der Forschung

Es gibt einige Verfahren, die sich mit der Rekonstruktion von Faserbündeln des menschlichen Gehirns aus DTI-Daten befassen.

Ein weit verbreiteter Ansatz ist das Fiber Tracking, das wiederum verschiedene Ausprägungen beinhaltet. Das wohl intuitivste Verfahren ist ein deterministisches, diskretes Tracking. Ausgangspunkt bilden dabei die Voxel einer Saatregion. Für jedes dieser Voxel wird jeweils das nächste untersucht, das in Richtung der Hauptdiffusionsrichtung des Ausgangsvoxels liegt. Werden dort die Abbruchkriterien erreicht, endet das Tracking, ansonsten wird die Hauptdiffusionsrichtung des Voxels als neue Faserrichtung genutzt. Zu den Abbruchkriterien zählen in erster Linie ein Schwellwert bzgl. der fraktionellen Anisotropie (FA) und bzgl. des Winkels zwischen der aktuellen und der vorangegangenen Richtung der Faser. Als weiteres Kriterium kommt oft noch eine minimale bzw. maximale Länge einer zu rekonstruierenden Faser hinzu [1, 2].

Das Ergebnis eines solchen Trackings liegt anschließend als Menge von einzelnen Faserlinien vor, wobei nicht jede Faser genau einer Faser des realen Bündels entspricht. In dieser Form erfüllt das Ergebnis jedoch noch nicht die Anforderungen für die Integration in die OP-Planung. Zu diesem Zweck müssen Hüllen erzeugt werden, die das rekonstruierte Bündel umschließen. Auch dazu gibt es verschiedene Ansätze. Die meisten beruhen darauf, entlang der Mittellinie des rekonstruierten Faserbahnen schrittweise begrenzende Kurven zu legen, die letztendlich durch Vernetzung ein Oberflächenmodell bilden [3]. Als begrenzende Kurven kommen zum Beispiel konvexe Hüllen in Betracht. Dazu werden entlang der Mittellinie Ebenen parallel zum Faserverlauf erzeugt und jeweils eine konvexe Hülle um die Schnittpunkte von rekonstruierten Fasern und der Ebene bestimmt [4].

## 3 Methoden

Die Grundidee der Methode besteht im Aufbau eines gerichteten und gewichteten Graphen, der durch einen minimalen Schnitt in zwei Bereiche aufgeteilt wird, die dann zum einen den Bereich des Faserbündels und zum anderen den Bereich außerhalb des Faserbündels darstellen.

### 3.1 Graphaufbau

Ausgehend von Saatregionen und Fiber Tracking wird eine initiale Mittellinie durch das zu segmentierende Bündel gelegt. Entlang dieser werden zu ihrem Verlauf orthogonale Ebenen $L_p$ $(p=0,...,P)$ erzeugt. Vom jeweiligen Durchstoßpunkt der Mittellinie werden äquidistant Strahlen $R_r$ $(r=0,...,R)$ ausgesendet und gleichmäßig an Punkten $P_i$ $(i=0,...,I)$ abgetastet, so dass insgesamt $(P+1)*(R+1)*(I+1)$ Evaluationspunkte entstehen [5,6,7]. Anhand der erzeugten Evaluationspunkte wird nun durch Entfaltung der Ebenen bzw. Strahlen ein gerichteter Graph $G = (V,E)$ aufgebaut (siehe Abbildung 1). Zunächst werden alle Evaluationspunkte als Knoten $v(p,r,i)$ (Evaluationspunkt $P_i$ in Ebene $L_p$, entlang des Strahls $R_r$) des Graphen hinzugefügt, sowie zwei weitere Knoten $s$ und $t$, die als Quelle und Senke fungieren. Die Knotenmenge $V$ ist dann bestimmt durch

$$V = \{v(p,r,i) | p \in [0,P], r \in [0,R], i \in [0,I]\} \cup \{s,t\}$$

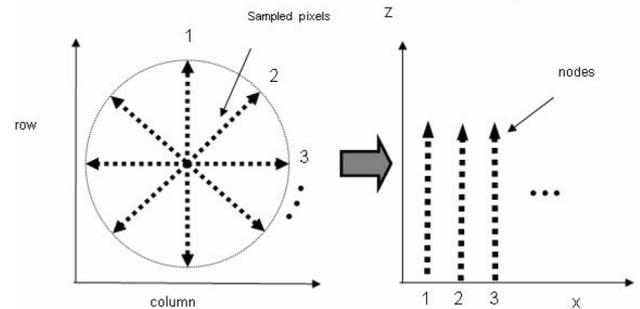

**Bild 1** Radiales Aussenden von Strahlen ausgehend von einem Centerlinepunkt in einer einzelnen Ebene (links) und Erzeugung eines entfalteten Bildes (rechts).

Zur Definition der Kantengewichte $w$ wird eine Kostenfunktion $c(p,r,i)$ genutzt. Für die mit $w$ gewichteten Kanten $(E = E_1 + E_2 + E_3 + E_{st})$ des Graphen werden verschiedene Schemata zur Definition angewendet.

1. ∞-gewichtete Kanten entlang eines Strahls:
   $E_1 = \{(v(p,r,i), v(p,r,i-1)) | i>0\}$
2. ∞-gewichtete Kanten zwischen Strahlen:
   $E_{21} = \{(v(p,r,i), v(p,r+1, max(0,i-\Delta_x)))\quad | r \in [0,R-1]\}$
   $E_{22} = \{(v(p,r,i), v(p,r-1, max(0,i-\Delta_x)))\quad | r \in [1,R]\}$
   $E_{23} = \{(v(p,0,i), v(p,R, max(0,i-\Delta_x)))\}$
   $E_{24} = \{(v(p,R,i), v(p,0, max(0,i-\Delta_x)))\}$
   $E_2 = E_{21} + E_{22} + E_{23} + E_{24}$
3. ∞-gewichtete Kanten zwischen Ebenen:
   $E_{31} = \{(v(p,r,i), v(p+1,r, max(i-\Delta_z)))\quad | p \in [0,P-1]\}$
   $E_{32} = \{(v(p,r,i), v(p-1,r, max(i-\Delta_z)))\quad | p \in [1,P]\}$
   $E_3 = E_{31} + E_{32}$
4. individuell gewichtete ($w$) Kanten zu $s$ und $t$:
   $E_{s1} = \{(v(p,r,0),s)\quad | w(p,r,i) = c(p,r,i)\}$
   $E_{t1} = \{(v(p,r,P),t)\quad | w(p,r,i) = c(p,r,i)\}$
   $E_{s2} = \{(v(p,r,i),s)\quad | i \in [1,P-1], c(p,r,i)-c(p,r,i-1)>0,$
   $\quad w(p,r,i) = |c(p,r,i)-c(p,r,i-1)|\}$
   $E_{t2} = \{(v(p,r,i),t)\quad | i \in [1,P-1], c(p,r,i)-c(p,r,i-1)<0,$
   $\quad w(p,r,i) = |c(p,r,i)-c(p,r,i-1)|\}$
   $E_{st} = E_{s1} + E_{s2} + E_{t1} + E_{e2}$

### 3.2 Kostenfunktion und Parameterwahl

Zur Definition der Kostenfunktion und somit für die Bestimmung eines optimalen s-t-Schnittes ist die Angabe ei-

nes individuellen mittleren Wertes der zu segmentierenden Struktur in den vorliegenden Daten notwendig. Die zugrundeliegenden Tensordaten können nicht direkt als ein einzelner Parameter genutzt werden. Jedoch lässt sich die fraktionelle Anisotropie als Segmentierparameter nutzen. Sie enthält zwar keine Richtungsinformationen, gibt aber durch den Betrag des Tensors, der der gerichteten Diffusion zugeschrieben wird, eine wichtige Auskunft über die Gerichtetheit der Diffusion und kann wie folgt aus den Eigenwerten ($\lambda_1$, $\lambda_2$, $\lambda_3$) des Diffusionstensors berechnet werden [8]

$$FA = \sqrt{\frac{(\lambda_1 - \lambda_2)^2 + (\lambda_2 - \lambda_3)^2 + (\lambda_1 - \lambda_3)^2}{2 \cdot (\lambda_1^2 + \lambda_2^2 + \lambda_3^2)}}$$

und liegt in einem Wertebereich von *0* (isotrope Diffusion) bis *1* (vollständig anisotrope Diffusion). Die spezifischen FA-Werte in der weißen Substanz können sich bei Patientendaten unterscheiden. Der FA-Wert variiert zum Beispiel altersbasiert, abhängig vom Entwicklungsstand [9]. Auch verschiedene Krankheiten wie z.B. M. Alzheimer können die FA-Werte in den Daten beeinflussen [10]. Durch diesen Umstand ist die Wahl eines jeweils angepassten mittleren FA-Wertes zur Definition der Kostenfunktion für jeden Datensatz notwendig.

Dazu kann man sich die abgetastete Mittellinie, die auch zum Graphaufbau genutzt wird, zu Nutze machen. Entlang dieser im Faserbündel liegenden Mittellinie können FA-Werte an den Abtastpunkten bestimmt und gemittelt werden. Um auch die Umgebung der Mittellinie einfließen zu lassen, kann für jeden Abtastpunkt anhand einer dreidimensionalen Maske aus der Umgebung ein mittlerer FA-Wert bestimmt werden, der wiederum mit den anderen so bestimmten Werten zu einem globalen Wert zusammengeführt wird. Ein Mittelwert *m* für einen Abtastpunkt an der Stelle (x,y,z) mit Intensitätswert I(x,y,z) im Datensatz wird dabei mit einer diskreten 3D-Maske (Nullpunkt im Zentrum der Maske) der Größe $m \times n \times l$ wie folgt berechnet

$$m = \sum_{i=-m/2}^{m/2} \sum_{j=-n/2}^{n/2} \sum_{k=-l/2}^{l/2} I(x+i, y+j, z+k) * mask(i,j,k)$$

Als mögliche Masken stehen Filter verschiedener Größen (*1×1×1*, *3×3×3*, bis *9×9×9*) zur Verfügung. Als Filter werden dabei ein Mittelwertfilter sowie ein diskreter Gaußfilter eingesetzt, wie sie exemplarisch für den zweidimensionalen Fall in Abbildung 2 dargestellt sind.

| 1/25 | 1/25 | 1/25 | 1/25 | 1/25 | | 1/256 | 4/256 | 6/256 | 4/256 | 1/256 |
|------|------|------|------|------|---|-------|-------|-------|-------|-------|
| 1/25 | 1/25 | 1/25 | 1/25 | 1/25 | | 4/256 | 16/256 | 24/256 | 16/256 | 4/256 |
| 1/25 | 1/25 | 1/25 | 1/25 | 1/25 | | 6/256 | 24/256 | 36/256 | 24/256 | 6/256 |
| 1/25 | 1/25 | 1/25 | 1/25 | 1/25 | | 4/256 | 16/256 | 24/256 | 16/256 | 4/256 |
| 1/25 | 1/25 | 1/25 | 1/25 | 1/25 | | 1/256 | 4/256 | 6/256 | 4/256 | 1/256 |

**Bild 2** 2D-Mittelwertfilter der Größe 5x5 (link), diskreter 2D-Gaußfilter der Größe 5x5 (rechts)

### 3.3 Min-Cut-Segmentierung

Nach dem Graphaufbau wir ein s-t-Schnitt in polynomialer Zeit berechnet, der eine optimale Segmentierung des Faserbündels erzeugt [11]. Durch die benutzerspezifischen Parameter $\Delta_x$ und $\Delta_y$ kann dabei noch die Glattheit und Steifigkeit der Segmentierung manuell beeinflusst werden.

### 3.4 Erzeugung eines 3D-Models

Der Min-Cut trennt die Knotenmenge in zwei disjunkte Mengen $V_1$ und $V_2$, die zum einen das Faserbündelinnere ($V_1$) und zum anderen das umliegende Gewebe ($V_2$) klassifizieren. Entlang jedes Strahls $R_r$ jeder Ebene $L_p$ kann nun ein Randpunkt $b(p,r)$ bestimmt werden durch

$$b(p,r) = v(p,r,i) + \frac{1}{2}(v(p,r,i+1) - v(p,r,i))$$

mit $v(p,r,i) \in V_1$ und $v(p,r,i+1) \in V_2$.
Die resultierende Punktwolke wird durch eine Triangulierung und anschließendes Voxelisieren in ein 3D Objekt umgewandelt.

## 4 Ergebnisse

Alle Methoden wurden in C++ innerhalb der Plattform MeVisLab (www.mevislab.de) implementiert. Die Berechnung des FA-Wertes, sowie die komplette Segmentierung benötigten nur wenige Sekunden (Intel Core 2 Quad CPU, 3GHz, 6 GB RAM, Windows XP Professional 2003, SP2).

Zur Evaluation des Ansatzes mit der automatisch bestimmten Kostenfunktion wurden zwei Software-Phantome genutzt, die Faserbündel mit bekanntem Verlauf und Ausmaß beinhalten: ein Torus-Phantom (Durchmesser der Querschnittsfläche: 10mm) und ein anatomisches Softwarephantom [12] mit modellierter rechter Pyramidenbahn. So besteht die Möglichkeit eines Vergleichs zwischen den Ergebnissen des Ansatzes und einer Referenz. Der Ansatz wurde ebenfalls auf Patientendaten angewendet, kann jedoch nur visuell evaluiert werden, da der Verlauf der Bahnen dort nicht a priori bekannt ist.

Zur Evaluation wird der Dice-Koeffizient (DSC), der üblicherweise in der medizinischen Bildverarbeitung zur Analyse der Überlappung zweier Segmentierungen A und B eingesetzt und wie folgt berechnet wird [13]

$$DSC = \frac{2 \cdot |A \cap B|}{|A| + |B|} \in [0,1]$$

Für das Torus-Phantom und auch das anatomische Softwarephantom wurde die Centerline jeweils an 50 bzw. 100 Stellen abgetastet. In jeder Untersuchungsebene wurde 30 Strahlen ausgesendet und an 30 Stellen (Abstand 0,5mm) Evaluationspunkte erzeugt. Abbildung 3 zeigt eine erzeugte Konturpunktwolke für die modellierte rechte Pyramidenbahn im anatomischen Softwarephantom in 3D-Ansicht und in einem axialen 2D-Schnittbild für einen Gaußfilter der Größe 3x3x3 bei 50 Evaluationsebenen.

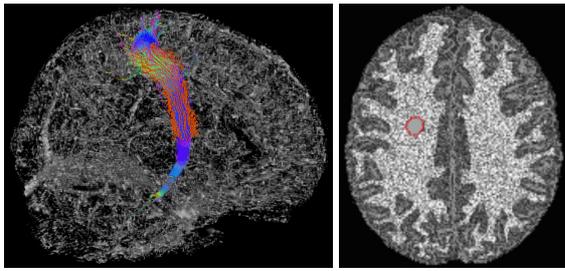

**Bild 3** Links: sagittale Ansicht der modellierten rechten Pyramidenbahn mit berechneter Kontur (rot), Rechts: axiale 2D-Schicht mit berechneter Kontur (rot).

Zum Vergleich der verschienen Masken zur automatischen Bestimmung der Kostenfunktion wurde ebenfalls eine manuelle Bestimmung der Kostenfunktion durchgeführt. Unter gleichen Parametereinstellungen ergab sich dabei für das Torus-Phantom ein mittlerer DSC von 74,46% ($\sigma = 0{,}026$). Für die modellierte Pyramidenbahn im anatomischen Software-Phantom konnte ein mittlerer DSC von 73,42% ($\sigma = 0{,}018$) mit der manuell bestimmter Kostenfunktion erreicht werden. Die DSCs der verschiedenen angewendeten Filter (homogen, Gauß) und Filtergrößen lieferten für beide Phantome ähnliche Werte wie beispielsweise in Tabelle 1 für das anatomische Phantom mit 50 Untersuchungsebenen aufgetragen ist.

| Mittelwert-filter | 1x1x1 | 3x3x3 | 5x5x5 | 7x7x7 | 9x9x9 |
|---|---|---|---|---|---|
| DSC (%) | 76,08 | 76,01 | 76,10 | 75,98 | 75,76 |
| **Gaußfilter** | 1x1x1 | 3x3x3 | 5x5x5 | 7x7x7 | 9x9x9 |
| DSC (%) | 76,08 | 76,01 | 76,09 | 76,13 | 76,38 |

**Tabelle 1** Ergebnisse der verschiedenen Filter für das Torus-Software-Phantom mit 100 Untersuchungsebenen.

Aus den Ergebnissen ist eine hohe Ähnlichkeit zwischen manueller Wahl und automatischer Bestimmung der Kostenfunktion erkennbar. Die verwendeten Filter unterscheiden sich dabei nicht signifikant für kleine Filtergrößen. Für größere Masken liefert jedoch der Gaußfilter bessere Ergebnisse. Bei der manuellen Wahl der Kostenfunktion ist eine visuelle Analyse der Faserumgebung notwendig, die mehrere Minuten in Anspruch nimmt. Durch die hier vorgeschlagene Automatisierung des Prozesses durch einen Filter, mit einer Berechnungszeit von 1-2 Sekunden, ist eine deutliche Beschleunigung möglich.

## 5 Zusammenfassung und Ausblick

In diesem Beitrag wurde ein neuer graph-basierter Ansatz zur Segmentierung von Bahnsystemen im menschlichen Gehirn vorgestellt. Der Ansatz nutzt zwei manuell gesetzte Saatregionen, die eine Auswahl der gewünschten Struktur liefern. Entlang der Centerline des getrackten Bündels werden in verschiedenen Ebenen gleich verteilte Strahlen ausgesendet und abgetastet. Die resultierende Menge an Evaluationspunkten wird zum Aufbau eines gerichteten und gewichteten Graphen genutzt. Dieser ist durch gewichtete Kanten mit einer Quelle und Senke verbunden. Durch einen Min-Cut wird eine optimale Segmentierung bezüglich der gegebenen Informationen erzeugt. Zur Evaluierung der Auswirkung verschiedener Kostenfunktionen wurden zwei Software-Phantome verwendet und über den DSC mit dem Verlauf und der Ausdehnung der dadurch vorgegebenen Faserbündel, sowie mit den Ergebnissen der manuellen Wahl der Kostenfunktion verglichen. In einem nächsten Schritt soll der Ansatz auch auf die Behandlung verzweigender Strukturen ausgedehnt werden. Außerdem soll der Ansatz durch möglichst automatische Platzierung der Saatregionen weiter automatisiert werden.

## 6 Literatur

[1] Mori, S.; van Zijl, P.C.M.: Fiber tracking: principles and strategies – a technical review. NMR Biomed. Vol. 15, 2002, pp. 468-480.

[2] Rueber, M.: Deterministic and probabilistic fiber tracking with diffusion tensor imaging under aversive diffusion conditions. Ph.D. thesis, Würzburg, 2009.

[3] Nimsky, Ch.; et al.: Visualization strategies for major white matter tracts identified by diffusion tensor imaging for intraoperative use. CARS, 2005, 793-797.

[4] Ding, Z.; et al.: Case study: reconstruction, visualization and quantification of neuronal fiber pathways. Proceedings of the conference on Visualization, 2001, pp. 453-456.

[5] Egger, J.; et al.: Aorta Segmentation for Stent Simulation. In: 12th Int. Conf. on MICCAI, Cardiovascular Interventional Imaging and Biophysical Modelling Workshop, London, UK, 2009.

[6] Egger, J.; et al.: Graph-Based Tracking Method for Aortic Thrombus Segmentation. In: 4th European Congress for Medical and Biomedical Engineering, Engineering for Health, Belgium, Springer, 2008.

[7] Li, K.; et al.: Optimal Surface Segmentation in Volumetric Images - A Graph Theoretic Approach. IEEE PAMI, 28(1): 119-134, 2006.

[8] Peeters, T.H.J.M.; et al.: Visualization and Processing of Tensor Fields, chap. Analysis of Distance/ Similarity Measures for Diffusion Tensor Imaging. Springer Berlin Heidelberg, 2009, pp. 113-136.

[9] van der Knaap, M.S.; et al.: Magnetic resonance of myelination and myelin disorders. Springer, 2005.

[10] Takahashi, S.; et al.: Selective reduction of diffusion anisotropy in white matter of Alzheimer disease brains measured by 3.0 Tesla magnetic resonance imaging. Neuroscience Letters 332, 45–48 (2002)

[11] Boykov, Y.; et al.: An Experimental Comparison of Min-Cut/Max-Flow Algorithms for Energy Minimization in Vision. IEEE PAMI, 26(9): 1124-1137, 2004.

[12] Barbieri, S.; et al.: Assessing Fiber Tracking Accuracy via Diffusion Tensor Software Models. SPIE, Vol. 7623, 2010.

[13] Zour, K. H.; et al.: Statistical Validation of Image Segmentation Quality Based on a Spatial Overlap Index: Scientific Reports, Academic Radiology, 11(2), pp. 178-189, 2004.